\pdfoutput=1

\documentclass[11pt]{article}

\usepackage[]{ACL2023}

\usepackage{times}
\usepackage{latexsym}
\usepackage{graphicx}
\usepackage{svg}
\usepackage{amsmath}
\usepackage{verbatim}
\usepackage{booktabs}
\usepackage{multirow}
\usepackage{amsfonts}
\usepackage{subcaption}
\usepackage[ruled, noend]{algorithm2e}
\usepackage{float}
\usepackage{textcomp}
\usepackage{makecell}
\usepackage{url}

\usepackage[T1]{fontenc}

\usepackage[utf8]{inputenc}

\usepackage{microtype}

\usepackage{inconsolata}

\title{Two is Better Than One: Answering Complex Questions by Multiple Knowledge Sources with Generalized Links}

\author{
Minhao Zhang$^1$\thanks{\hspace{0.15cm}Minhao Zhang contributes this work during his internship at Langboat.}\quad Yongliang Ma$^2$\quad Yanzeng Li$^1$\quad Ruoyu Zhang$^1$ \\ \bf Lei Zou$^1$\thanks{\hspace{0.15cm}Corresponding Author}\quad Ming Zhou$^2$ \\
$^1$Wangxuan Institute of Computer Technology, Peking University. Beijing, China \\
$^2$Langboat Technology, Beijing, China \\
$^1$\texttt{\{zhangminhao, ry\_zhang, zoulei\}@pku.edu.cn, liyanzeng@stu.pku.edu.cn} \\
$^2$\texttt{\{mayongliang, zhouming\}@langboat.com}
}

\begin{document}
\maketitle
\begin{abstract}
Incorporating multiple knowledge sources is proven to be beneficial for answering complex factoid questions. To utilize multiple knowledge bases (KB), previous works merge all KBs into a single graph via entity alignment and reduce the problem to question-answering (QA) over the fused KB. In reality, various link relations between KBs might be adopted in QA over multi-KBs. In addition to the identity between the alignable entities (i.e. full link), unalignable entities expressing the different aspects or types of an abstract concept may also be treated identical in a question (i.e. partial link). 
Hence, the KB fusion in prior works fails to represent all types of links, restricting their ability to comprehend multi-KBs for QA.
In this work, we formulate the novel Multi-KB-QA task that leverages the full and partial links among multiple KBs to derive correct answers, a benchmark with diversified link and query types is also constructed to efficiently evaluate Multi-KB-QA performance. Finally, we propose a method for Multi-KB-QA that encodes all link relations in the KB embedding to score and rank candidate answers. Experiments show that our method markedly surpasses conventional KB-QA systems in Multi-KB-QA, justifying the necessity of devising this task. 

\end{abstract}

\section{Introduction}
\label{sec:intro}

Knowledge base question-answering (KB-QA) consults fact triples stored in the background knowledge base (KB) to answer factoid questions \citep{unger2012template}. 
Despite the success in simple (one-hop) questions, early attempts \citep{bordes2015large, petrochuk-zettlemoyer-2018-simplequestions} fall short in solving complex questions that require different knowledge sources to accurately retrieve the answers. Hence, several systems leverage auxiliary documents, tables, or even images to supplement a KB in complex QA scenarios \citep{lv2020graph, shah2019kvqa}. Meanwhile, a separate line of investigations combine multiple KBs to tackle complex questions. Specifically, multiple KBs are fused into a single graph via entity alignment \citep{zeng2021comprehensive} and the KB-QA system executes on such graph to access facts of different origins \citep{song2015tr, zhang2016joint, luo2022mu}.

\begin{figure*}[t]
    \centering
    \includegraphics[width=0.84\linewidth]{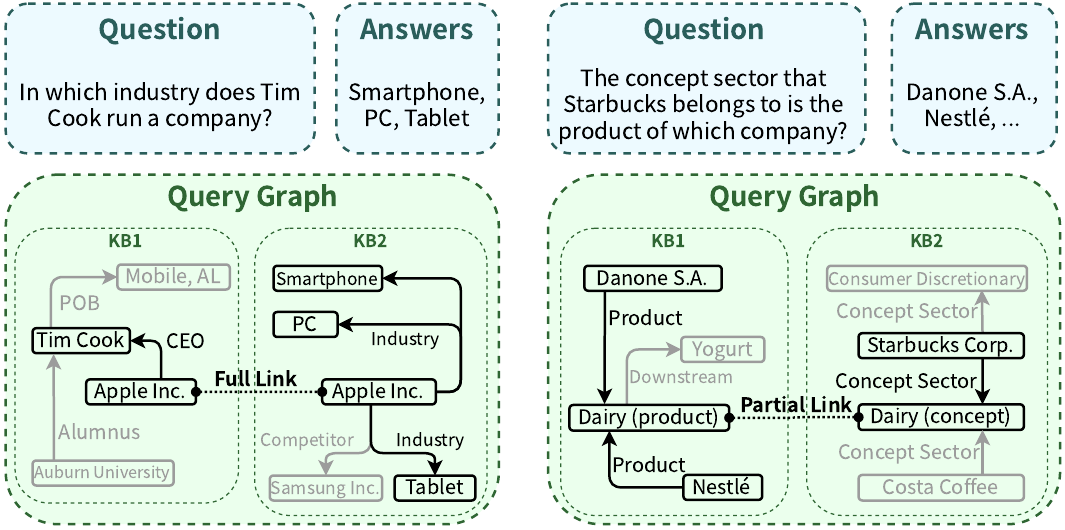}
    \caption{Leveraging full and partial links between multiple KBs to answer complex questions. Both questions require certain facts from each KB to accurately select answers, further, understanding the linking points between KBs is also essential. For the left case, the entity \textit{Apple Inc.} from both KBs refer to the same company (i.e. a \textbf{full link} exists between them), serving as the link to integrate the facts from both sides and to answer the left question. For the right case, though the entity \textit{Dairy (product)} and \textit{Dairy (concept)} are distinct in semantics, they might be considered as one in a casual QA scenario (i.e. a \textbf{partial link} exists between them). To this regard, a KB-QA system should grasp such partial links to generate satisfying answers even though the entities cannot be directly aligned.}
    \label{fig:full_partial_link}
\end{figure*}

Directly merging multiple KBs does enable a system to seek better answers from multiple knowledge sources, but it appears insufficient in fully representing the diversified link relationships between different KBs, which hinders the QA system to grasp and utilize all the knowledge implied by multiple KBs. More concretely, as shown in Figure \ref{fig:full_partial_link}, two types of intrinsic links (namely full and partial link) between multi-KBs are used in answering complex questions. For the left case, the question assumes that the entity \textit{Apple Inc.} in both KBs are identical.
Directly linking identical entities on both sides resembles merging aligned entities as in the aforementioned works, we name such identity as \textbf{full links}. 
Nevertheless, questions on multi-KBs may also involve other inter-KB relations beyond full links. As in the right case of Figure \ref{fig:full_partial_link}, the \textit{Concept Sector}\footnote{\textit{Concept Sector} refers to a collection of stocks in a particular field, see \url{https://www.globexmarkets.com/investment-sectors.html} for detailed explanations.} of \textit{Starbucks Corp.} is first queried to be \textit{Dairy (concept)} in KB2, which is then regarded the same as the entity \textit{Dairy (product)} in KB1 despite their discrepancy in semantics, finally, the answers are derived from the neighbors of \textit{Dairy (product)} in KB1. In conventional entity alignment, \textit{Dairy (concept)} and \textit{Dairy (product)} refer to different facts and are hence unlinkable, yet, owing to the casualness of natural language, they should be treated as identical in certain questions and we name such identity as \textbf{partial links}. In this regard, previous attempts to merge multiple KBs by entity alignment cannot wholly exploit the generalized links among KBs (with both full and partial links), limiting their capability in answering questions that embody varied cross-KB relations (see Appendix \ref{sec:partial_failure} for more discussions on the deficiency of entity alignment over partial links).

In this paper, we study the task of question-answering over multiple KBs with generalized links (Multi-KB-QA), starting by formulating the full and partial links between KBs and shedding light on the way such links are employed in KB-QA. Besides, former KB-QA benchmarks are shown to be answerable by a single KB \citep{yih2016value, trivedi2017lc}, impeding it from effectively evaluating the ability to leverage generalized links for QA. Thus, we construct a new benchmark consisting of human-curated QA-pairs based on various full and partial links over two large-scale KBs. Statistics are also provided to show the diversity and completeness of the benchmark w.r.t. link type and query structure. With this benchmark, we propose a simple yet effective baseline approach inspired by \citet{saxena2020improving} to reveal the value of formulating the Multi-KB-QA task. Specifically, we train a link-aware embedding for each KB powered by a translator module that grasps the semantics of all generalized links in embedding space. With all link relations encoded, we apply such embedding to score each candidate answer w.r.t. the question and the topic entity, forming the final predictions of our system. Notwithstanding its simplicity, experiments show that our approach apparently outperforms prior state-of-the-art methods that fuse multi-KBs, demonstrating the necessity of incorporating generalized links among multi-KBs in KB-QA. Additional analysis also exhibit the inherent pluggability of Multi-KB-QA to optimize its efficiency in practical use.

In short, the contributions of this paper are: 1) we formulate and illustrate the novel Multi-KB-QA task centering on the generalized links between KBs; 2) we publish a comprehensive and diversified benchmark on Multi-KB-QA to prompt future study; 3) we devise a baseline Multi-KB-QA method to comprehend and employ all generalize links, by surpassing conventional KB-QA methods, we justify the value of Multi-KB-QA.

\section{Formulation of Multi-KB-QA}
\label{sec:formulation}
We describe the Multi-KB-QA task in this section.
\subsection{Preliminaries}
A KB on an entity set $E$ contains knowledge triples to represent the relations between entities ($KB = \{<s,r,o>|s,o\in E\}$).
As in normal factoid QA settings \citep{berant2013semantic}, Multi-KB-QA overall asks systems to answer a natural language question $q$ by providing a set of facts from the KB ($A \subseteq E$).

Meanwhile, each question is intrinsically associated with a query graph $Q\subseteq KB$ that embodies the reasoning path in the question (black lines represent the query graph in Figure \ref{fig:full_partial_link}; e.g. the left question queries the relation path \textit{CEO-Industry} around the entity \textit{Tim Cook}, thus, the subgraph corresponding to this path in KB constitutes the query graph). Since the query graph denotes the intention of the question and the answers are exactly part of the nodes in this graph, it is viewed as a vital bridge between question and answer in both constructing KB-QA datasets \citep{gu2021beyond} and performing KB-QA \citep{zhang2022crake}.

\subsection{QA on Multiple KBs}
As stated in Section \ref{sec:intro}, Multi-KB-QA consider the scenario where facts from multiple KBs (say $KB_1$ and $KB_2$ on entity set $E_1$ and $E_2$) are essential to obtain answers. In this case, the query graph of each question could be divided into multiple separate subgraphs corresponding to the reasoning path in each KB ($Q=Q_1\cup Q_2$, where $Q_1\subseteq KB_1,Q_2\subseteq KB_2$). Back to the query graph in Figure \ref{fig:full_partial_link}-left, it contains a subgraph in KB1 (that stores the \textit{CEO} of \textit{Apple Inc.}) and KB2 (that stores the \textit{Industry} of \textit{Apple Inc.}) respectively.

\subsection{Generalized Links Between KBs}
\label{sec:form_gen_links}
Defining multiple sub-query-graphs for each KB is not enough, only when all subgraphs are linked together can they fully express the intention of a question. As discussed in Section \ref{sec:intro}, such links occur between the entities from different KBs that are treated as identical in the question, which can further be classified into full and partial links. Here, we explain these generalized links in more detail.
\paragraph{Full Links} When a pair of entities from different KBs instantiate the same object, we view that a full link exists between them. 
For instance, a full link exists between \textit{Apple Inc.} in KB 1 and 2 in Figure \ref{fig:full_partial_link}-left. 
In entity alignment, entities with such links can be directly aligned to merge multiple KBs.
\paragraph{Partial Links} A pair of entities without a full link does not entail their distinguishment in a query graph. Instead, entities that express the different aspects, stages, or types of an abstract concept may also be referred as identical in some queries. In Figure \ref{fig:full_partial_link}-right, the entity \textit{Dairy (product)} in KB1 instantiates the abstract concept "dairy" by denoting the products made of milk, while the entity \textit{Dairy (concept)} in KB2 instantiates "dairy" as the concept sector of dairy in stock market; despite the difference in meaning, they actually serve as the joint point between the two sub-query-graphs to express the whole question. In addition to this example, the identity between an entity pair without a full link
may appear in various formats (e.g. between a stock sector and a company, between a football club at different years, etc.) and we conclude it as partial links. Disparate with full links, the partially-linked entities cannot be fused in entity alignment since they represent distinct instances, their oneness in the query graph results merely from the casualness in the question that makes no guarantee on a rigorous query graph linkage. Please refer to Appendix \ref{sec:partial_failure} for a detailed analysis of partial links and the value of formulating Multi-KB-QA.

Combining full and partial links, we can now define the generalized links between $KB_1$ and $KB_2$ by $L=\{<e_1, e_2, t>|e_1\in E_1, e_2\in E_2, t\in \{0,1\}\}$ where $t$ denotes the type of each link (0/1 for full/partial link respectively). Eventually, we conclude that Multi-KB-QA diverge from normal KB-QA in that the query graph of each question has multiple subgraphs belonging to multi-KBs while all the subgraphs are correlated by generalized links. To this end, Multi-KB-QA tests the ability to integrate multiple knowledge sources by generalized links in answering complex questions.

\begin{table}[t]
\centering

\resizebox{0.99\columnwidth}{!}{
\begin{tabular}{l l}
    \toprule
    \bfseries Type & \bfseries Example\\
    \cmidrule(lr){1-2}
    CS - Company & Alibaba\_(concept) - Alibaba\_Group \\
    CS - Product & OLED - OLED\_Screen \\
    CS - Industry & 5G\_(concept) - 5G \\
    CS - Person & Yun\_Ma\_(concept) - Jack\_Yun\_Ma \\
    SC - Industry & IC\_Manufacturing - Microchip \\
    \bottomrule
\end{tabular}}
\caption{Examples of the extracted partial links. \textit{Type} denotes the type of the partially-linked entities in both KBs. \textit{CS} and \textit{SC} abbreviates the entity type \textit{Concept\_Sector} and \textit{Supply\_Chain} respectively. We translate the raw Chinese entity names for presentation.}

\label{tab:partial_link_type}
\end{table}

\section{The MKBQA Dataset}
\label{sec:dataset}
To evaluate against the setting of Multi-KB-QA, the questions should satisfy the features introduced in Section \ref{sec:formulation}, i.e. utilize multi-KBs and generalized links. However, prior datasets in KB-QA are either answerable barely by a single KB \citep{trivedi2017lc} or devised to work on a large KB composed by several sub-KBs while all sub-KBs are linked beforehand by predefined full links \citep{ngomo20189th}. Hence, we construct a new dataset to effectively benchmark Multi-KB-QA.

\subsection{Extracting Generalized Links}
\label{sec:extract_gen_links}

The core of building a Multi-KB-QA benchmark is the discovery of generalized links between KBs. Section \ref{sec:formulation} hints the potential existence of various partial links among financial concepts, therefore, we build our dataset upon two large-scale finance-domain KBs curated from financial news, research reports and company announcements, mainly focusing on stock sectors and company information respectively. To mine generalized links between them, we first match similar entities between two KBs based on Levenshtein distance \citep{Soukoreff2001MeasuringEI} and treat matched entity pairs with the same type as full links, leaving others as partial links. Although such method cannot exhaust all possible links between KBs, with a high similarity threshold, it can achieve high linking accuracy while remaining enough amount of links to establish a high-quality KB-QA dataset upon them. Altogether, we obtain 4452 pairs of links with 5 distinct types of partial links as shown in Table \ref{tab:partial_link_type}.

\begin{figure}[t]
    \centering
    \includegraphics[width=0.82\linewidth]{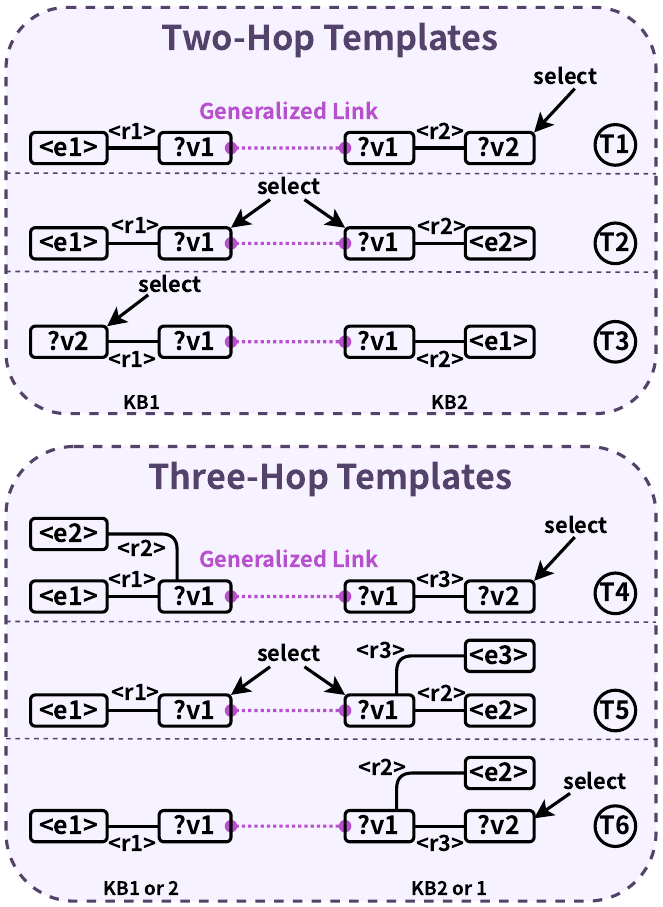}
    \caption{
    Six templates are used to generate Multi-KB-QA query graphs. To instantiate a template, we sample subgraphs in the KBs that match the pattern of the template (i.e. each \textit{<e>} node and \textit{<r>} edge corresponded by an entity and relation respectively). See example query graphs in Appendix \ref{sec:example_data}.
    }
    \label{fig:query_graph}
\end{figure}

\subsection{Constructing KB-QA Dataset}
\label{sec:construction_dataset}
Based on the discovered links, we construct our dataset following the common process for KB-QA dataset construction in \citet{Wang2015BuildingAS}, in which the query graph is first sampled from the KB and then automatically verbalized as a canonical question by templates, the question is finally paraphrased by human annotators to improve naturalness and diversity.

To incorporate generalized links into complex questions, we devise several query graph templates in Figure \ref{fig:query_graph}, by filling in certain entities and relations from KB, multiple query graphs can be instantiated from each template. After forming a query graph, we obtain its canonical question by rules and derive its answers from the KB (unlike normal query graphs, our graph on multi-KBs cannot be directly converted to a KB query to get answers, we hence separately acquire the results for each hop and manually synthesize the answers).

Finally, since the canonical question is largely rule-based, we ask annotators to give each question a natural paraphrase in Chinese while persisting its semantics, i.e. remaining accord with the query graph. Specifically, annotators with professional KB-QA experience are hired and are provided a detailed training to improve the quality and diversity of the paraphrase. Next, the first round of annotation is conducted on 5\% of the data and the results are discussed by all annotators to improve coherence. Lastly, all the rest data are paraphrased and all the question-answer pairs are now formed to be our \textbf{MKBQA} benchmark.

\begin{table}[t]
\centering
\resizebox{0.9\columnwidth}{!}{
\begin{tabular}{l c c c c c c}
    \toprule
    \multirow{2}{*}{\bfseries Template} & 
    \multicolumn{3}{c}{\bfseries Two-Hop} &
    \multicolumn{3}{c}{\bfseries Three-Hop}\\
    \cmidrule(lr){2-4}\cmidrule(lr){5-7} & T1 & T2 & T3 & T4 & T5 & T6\\
    \cmidrule(lr){1-7}
    Full Link & 108 & 576 & 429 & 68 & 105 & 101\\
    Partial Link & 297 & 415 & 243 & 0 & 0 & 0\\
    \bottomrule
\end{tabular}}
\caption{The number of data points of each link type and query graph structure in the MKBQA benchmark.}
\label{tab:stats}
\end{table}

\subsection{Data Statistics}
We provide the breakdown of question type in Table \ref{tab:stats} for reference, demonstrating that our benchmark covers both full and partial links while attaining diversity in query structure.

\section{Baseline Method for Multi-KB-QA}
\label{sec:method}

\begin{figure*}[t]
    \centering
    \includegraphics[width=0.99\linewidth]{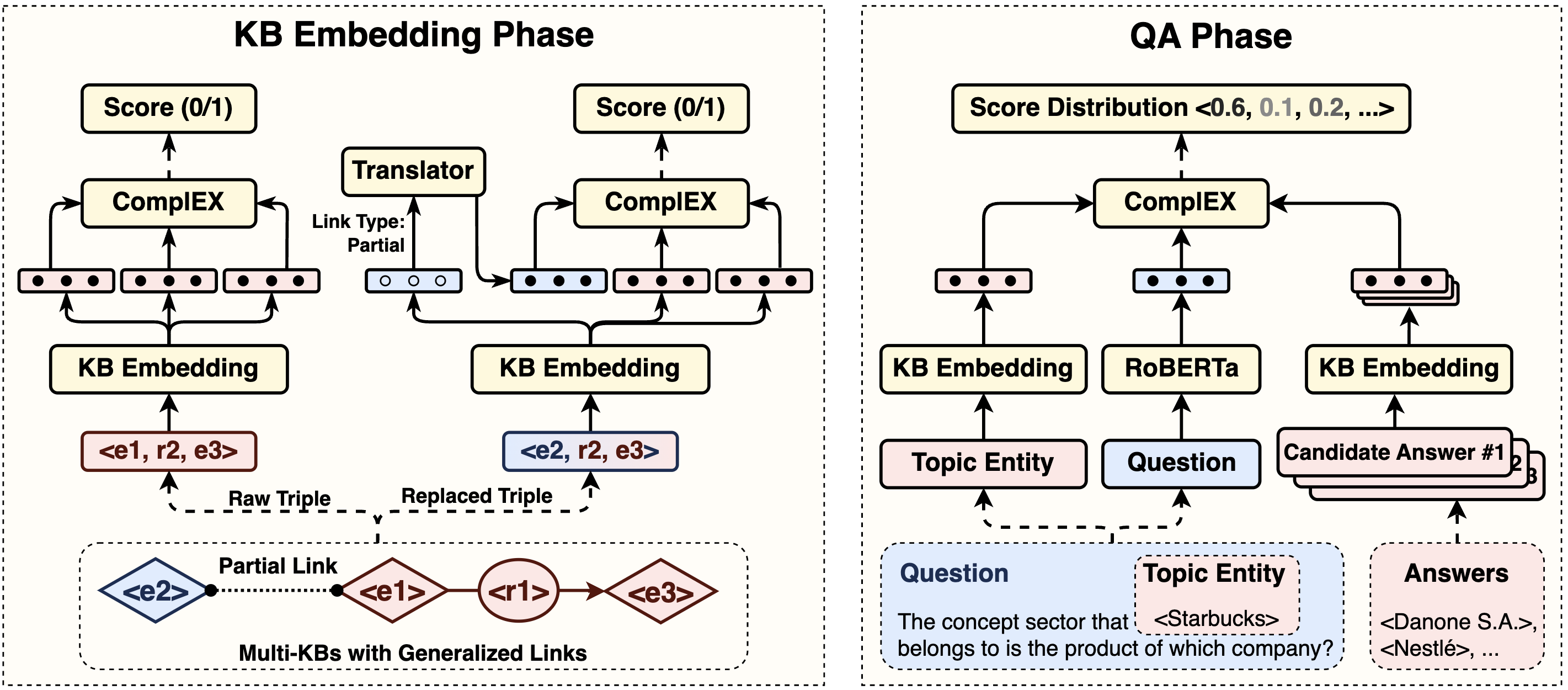}
    \caption{The overview of the proposed baseline Multi-KB-QA method.}
    \label{fig:method_overview}
\end{figure*}

To unveil the value of Multi-KB-QA, we need to show that existing KB-QA methods cannot yield satisfying results in such task. As discussed in Section \ref{sec:intro}, most prior works tend to fuse multi-KBs by entity alignment, failing to take advantage of the unalignable partial links. Therefore, we devise a baseline approach to leave each KB separate while encoding all generalized links. Although prior works excel in normal KB-QA, our simple yet effective baseline evidently outperforms them in the Multi-KB-QA setting (Section \ref{sec:exp}), indicating that Multi-KB-QA cannot be effectively solved by conventional KB-QA systems.

This section, we present our baseline method that first trains a KB embedding to encode generalized links (Section \ref{sec:kb_emb_phase}) and then uses these vectors to score and rank candidate answers (Section \ref{sec:qa_phase}).

\subsection{KB Embedding Phase}
\label{sec:kb_emb_phase}
Directly injecting generalized links to a KB-QA model is not trivial, however, efforts are made to jointly encode multiple KBs with full links \citep{trivedi-etal-2018-linknbed}, prompting us to first encode generalized links in the KB embedding and employ such embedding (instead of the links) in QA afterwards.

The overview of our embedding approach is shown in Figure \ref{fig:method_overview}-left. Like normal KB embedding methods \citep{Bordes2013TranslatingEF, balazevic-etal-2019-tucker}, we basically adopt a function to score triples and train the embedding vectors of each entity and relation such that the score of the triples present in the KB is higher than the negative samples. Specifically, for each triple $\text{<}s,r,o\text{>}\in KB$, where $s,r,o\in \mathbb{N}$ denotes the index of each entity and relation, we obtain its $h$-dimensional embedding vectors by $Emb:\mathbb{N}\rightarrow \mathbb{C}^{h}$
\[ \mathbf{h_{\{s,r,o\}}} = Emb(\{s,r,o\}) \in \mathbb{C}^h \]
and use the $\text{ComplEx}:\mathbb{C}^h\times\mathbb{C}^h\times\mathbb{C}^h\rightarrow \mathbb{R}$ function \citep{Trouillon2016ComplexEF} to score the triple.
\[ \lambda_{s,r,o} = \text{Sigmoid}(\text{ComplEx}(\mathbf{h_s},\mathbf{h_r},\mathbf{h_o}))\in \mathbb{R} \]
\[ \text{ComplEx}(\mathbf{h_1},\mathbf{h_2},\mathbf{h_3}) := \texttt{Re}(\sum\limits_{i=1}^h \mathbf{h}_1^{(i)}\mathbf{h}_2^{(i)}\mathbf{\Bar{h}}_3^{(i)}) \]
To train the embedding, we randomly collect k negative samples $\{\Tilde{o}_i\}$ for $o$ (i.e. $\forall i, \text{<}s,r,\Tilde{o}_i\text{>}\notin KB$) and compute the constrastive loss $\ell_{raw}$.
\[\ \ell_{raw} = \sum\limits_{\text{<}s,r,o\text{>}\in \text{KB}} -log(\lambda_{s,r,o}) - \sum\limits_{i=1}^k log(1-\lambda_{s,r,\Tilde{o}_i}) \]

This common process learns the entity semantics inside each KB but omits the relations between KBs, we devise the following process to encode the links into the embedding as well. As mentioned in Section \ref{sec:form_gen_links} and \ref{sec:extract_gen_links}, each link can be expressed by a pair of entities $e_1\in E_1, e_2\in E_2$ from KB 1 and 2 and the type of the link $t\in \{0,1\}$, forming the link set $L$. For each raw triple, we replace its subject if it appears in the link set to form the replace set of two directions $R_{\{12,21\}} = \text{\{<} \Hat{s},r,o,t \text{>\}}|\text{<}s,r,o\text{>}\in\ KB_{\{1,2\}},\text{<} \text{\{}s,\Hat{s}\text{\}},\text{\{}\Hat{s},s\text{\}},t \text{>}\in L\text{\}}$. To encode the link relations, we could conduct the similar triple-scoring and contrastive-training on $R_{12}$ and $R_{21}$, but the entity $\Hat{s}$ and $o$ in the replace set originate from different KBs with disparate semantic spaces, directly scoring the replaced triple may hence be confusing for the embedding. Thus, we adopt a translator to bridge the semantic spaces. Concretely, we introduce a vector for each link type by $E_t:\text{\{}0,1\text{\}}\rightarrow \mathbb{C}^h$ and use a FCN $Trans: \mathbb{C}^{2h}\rightarrow \mathbb{C}^h$ to shift $\Hat{s}$ to the semantic space of $s$ in compliance with $r$ and $o$ ($\oplus$ denotes concatenation).
\[ \Tilde{\mathbf{h}}_{\Hat{s}} = Trans(\mathbf{h}_{\Hat{s}}\oplus E_t(t)) \in \mathbb{C}^h \]
Note that the type of the link may affect the difference between $s$ and $\Hat{s}$ (e.g. a full link may entail a relatively close vector representation, while partially-linked entities may have disparate representation w.r.t. type or aspect), so $t$ is also engaged in the translation. Now, we can properly score the replaced triple by $\lambda_{\Hat{s},r,o}$ and encode all the generalized links to the embedding by minimizing $\ell_{link}$ on $R=R_{12}\cup R_{21}$.
\[ \lambda_{\Hat{s},r,o} = \text{Sigmoid}(\text{ComplEx}(\mathbf{\Tilde{h}_{\Hat{s}}},\mathbf{h_r},\mathbf{h_o}))\in \mathbb{R} \]
\[\ \ell_{link} = \sum\limits_{\text{<}\Hat{s},r,o\text{>}\in R} -log(\lambda_{\Hat{s},r,o}) - \sum\limits_{i=1}^k log(1-\lambda_{\Hat{s},r,\Tilde{o}_i}) \]
Finally, we minimize the loss $\ell =\ell_{raw}+r_{lk}\ell_{link}$ to obtain a link-aware embedding for each entity.

\subsection{QA Phase}
\label{sec:qa_phase}
Inspired by \citet{saxena2020improving}, we directly exploit the KB embedding to retrieve answers as shown in Figure \ref{fig:method_overview}-right. For each question $\mathbf{q}\in \mathbb{N}^n$, we use a RoBERTa \citep{Liu2019RoBERTaAR} encoder $E_{rb}:\mathbb{N}^n\rightarrow \mathbb{R}^d$ and a FCN\footnote{In practice, the FCN converts the input to a 2h-dimensional real vector and its first and second half is viewed as the real and imaginary part of the output $\mathbb{C}^h$ vector.} $R2C:\mathbb{R}^d\rightarrow\mathbb{C}^n$ to encode it into a complex vector of the same length with the KB embedding.
\[ \mathbf{h_{q}} = R2C(E_{rb}(\mathbf{q})) \in \mathbb{C}^h \]
For each topic entity $\text{\{}e_i\text{\}}\subseteq \mathbb{N}$ and answer $\text{\{}a_j\text{\}}\subseteq \mathbb{N}$ of $\mathbf{q}$, we obtain their embedding $\mathbf{h_{e_i}}, \mathbf{h_{a_j}}\in\mathbb{C}^h$ as in Section \ref{sec:kb_emb_phase} and compute the likelihood that $a_j$ answers $\mathbf{q}$ by:
\[ \lambda_{e_i, a_j}=\text{Sigmoid}(\text{ComplEx}(\mathbf{h_{e_i}},\mathbf{h_q},\mathbf{h_{a_j}}))\in \mathbb{R} \]
In prediction, we average the score given by each topic entity $\lambda_{a_j}=\sum\limits_i\lambda_{e_i,a_j}/|\text{\{}e_i\text{\}}|$ to score and rank the whole entity set $E$ and output the top-scored answers. To train the QA model, we sample incorrect answers ${\Tilde{a}_k}$ for each $\mathbf{q}$ and minimize $\ell_q$.
\[ \ell_{q} = -\sum\limits_{i,j}log(\lambda_{e_i,a_j})-\sum\limits_{i,k}log(1-\lambda_{e_i,\Tilde{a}_k}) \]

With this scheme, the QA model can learn to utilize the generalized links encoded in the embeddings in comprehending Multi-KB-QA questions.

\subsection{Pluggable Training and Inferencing}
\label{sec:pluggability}
In Multi-KB-QA, all KBs are treated separate (not fused to a single KB), granting the methods for this task an intrinsic pluggability, i.e. we can perform QA on a single KB when no other KBs are present, while an extra KB can also be naturally plugged to the system without re-training on all KBs.

Concretely, we first train on $KB_1$ to obtain its embedding, to plug in another KB, we simply train the embedding of the new entities on $KB_2$ and $R_{21}$ while fixing the embedding of KB1. In practical use where multi-KBs are not simultaneously obtained, such pluggability can notably reduce the training cost when adding a new KB to our system.
Besides, at inference, we can also use a specific KB to derive answers instead of searching all KBs to improve efficiency (i.e. unplug a KB).

\begin{table}[t]
\centering
\resizebox{1.0\columnwidth}{!}{
\begin{tabular}{l c c c c}
    \toprule
    \multirow{2}{*}{\bfseries Method} & 
    \multicolumn{2}{c}{\bfseries Dev Set} &
    \multicolumn{2}{c}{\bfseries Test Set}\\
    \cmidrule(lr){2-3}\cmidrule(lr){4-5} & MRR & Hits@1 & MRR & Hits@1\\
    \cmidrule(lr){1-5}
    No-Link & .192 \textpm\space.000 & .113 \textpm\space.011 & .166 \textpm\space.010 & .088 \textpm\space.011\\
    Merge-KB & .351 \textpm\space.004 & .265 \textpm\space.011 & .350 \textpm\space.011 & .242 \textpm\space.011\\
    \cmidrule(lr){1-5}
    Full-Link & .371 \textpm\space.011 & .261 \textpm\space.009 & .374 \textpm\space.021 & .254 \textpm\space.022\\
    Multi-KB & \bfseries.494 \textpm\space.003 & \bfseries.373 \textpm\space.007 & \bfseries.488 \textpm\space.017 & \bfseries.363 \textpm\space.021\\
    \bottomrule
\end{tabular}}
\caption{The overall performance on the MKBQA benchmark. \textbf{No-Link} and \textbf{Merge-KB} are conventional KB-QA baselines mentioned in Section \ref{sec:exp}. \textbf{Multi-KB} refers to our method in Section \ref{sec:method}. We also evaluate our method while excluding all partial links when training the embedding 
in \textbf{Full-Link} to reveal the effectiveness of our approach and the value of generalized links.}
\label{tab:overall_eval}
\end{table}

\begin{table*}[t]
\centering
\resizebox{0.70\linewidth}{!}{
\begin{tabular}{l c c c c c c c c c}
    \toprule
    \multirow{2}{*}{\bfseries Method} & 
    \multicolumn{6}{c}{\bfseries Full Link} &
    \multicolumn{3}{c}{\bfseries Partial Link}\\
    \cmidrule(lr){2-7}\cmidrule(lr){8-10} & T1 & T2 & T3 & T4 & T5 & T6 & T1 & T2 & T3\\
    \cmidrule(lr){1-10}
    No-Link & .198 & .044 & .313 & .260 & .004 & .050 & .195 & .238 & .357\\
    Merge-KB & \bfseries.632 & .321 & \bfseries.639 & .291 & .122 & \bfseries.439 & .151 & .265 & .313\\
    \cmidrule(lr){1-10}
    Full-Link & .597 & .442 & .480 & .275 & .313 & .287 & .236 & .254 & .425\\
    Multi-KB & .441 & \bfseries.473 & .505 & \bfseries.426 & \bfseries.335 & .314 & \bfseries.337 & \bfseries.702 & \bfseries.557\\
    \bottomrule
\end{tabular}}
\caption{The MRR on dev set broken-down to different link types and query structures. See Table \ref{tab:overall_eval} and Figure \ref{fig:query_graph} for the definition of each method and query structure (T1$\sim$T6 on full and partial links).}
\label{tab:breakdown_eval}
\end{table*}

\section{Experiments}
\label{sec:exp}
\paragraph{Dataset} We conduct experiments on the MKBQA benchmark detailed in Section \ref{sec:dataset} (see Appendix \ref{sec:appendix_anno} for the annotation procedure). For experiment, we randomly split the annotated data into the train/dev/test set with 1873/235/234 questions respectively. The benchmark is released with a GPL-3.0 licence to promote future studies focusing on answering the questions over multi-KBs that involve generalized links. The benchmark is evaluated on the average MRR \citep{Craswell2009MeanRR} of the answers (the highest rank is counted when there are multiple answers) while Hits@1 is also provided to show the accuracy of the first predicted answer.

\paragraph{Baselines} We compare our approach against conventional KB-QA methods adapted to the setting of Multi-KB-QA, two baselines are concluded from prior works: 1) \textbf{No-Link} treats each KB as a separate knowledge source and ignores the links between KBs \citep{fader2014open}, in practice we embed each KB separately and train the QA model based on such embedding to compare with our approach. 2) \textbf{Merge-KB} fuses all KBs to a single graph by full links \citep{zhang2016joint, luo2022mu} and achieves SoTA accuracy on conventional KB-QA datases like WebQuestionsSP \citep{yih2016value}, we train the QA model on the embedding of the fused graph in practice. All these methods fail to utilize all generalized links among KBs.

\paragraph{Setup} We apply the ext-large version of Chinese RoBERTa \citep{cui-etal-2021-pretrain} as our question encoder. Experiment results are averaged over 3 runs on an RTX 3090 GPU. The KB embedding is acquired by 400 iterations over all triples ($\sim$72 hrs) and we save the checkpoint with best the training MRR. The QA model is trained for 1000 epochs ($\sim$20 hrs) while fixing the KB embedding and the checkpoint of the highest MRR on dev set is reported. We mainly follow the hyperparameter settings in \citet{saxena2020improving} without additional hyperparameter search, see Appendix \ref{sec:appendix_hyper} for the detailed settings.

\subsection{Overall Evaluation}
\label{sec:overall_exp}
The overall performance of the conventional KB-QA baselines and our approach is illustrated in Table \ref{tab:overall_eval}. Though both No-Link and Merge-KB utilizes multiple KBs to provide answers, the latter leverages the full links to merge KBs and clearly surpasses the former, indicating that full links does help the systems in comprehending and using multiple knowledge sources for QA. Such results conform with the superior performance of Merge-KB methods like \citet{luo2022mu} on convertional KB-QA. 
However, our approach (Multi-KB) consistently outperforms Merge-KB on MKBQA by a large margin ($\sim$12\% gain on all metrics). To understand the gap, we obtain a KB embedding by the same process in Section \ref{sec:kb_emb_phase} except that only full links are employed in training and evaluate its QA results in Table \ref{tab:overall_eval} (Full-Link).

In general, Multi-KB differs from Merge-KB in two aspects: 1) it leaves multiple KBs separate in embedding to avoid fusing all KBs to a single one; 2) by circumventing the entity alignment in graph fusion, it is capable to encode the unalignable partial links to assist QA. Clearly, Full-Link differs from Merge-KB only by the first point. As in Table \ref{tab:overall_eval}, Full-Link slightly outperforms Merge-KB to imply the effectiveness of our KB embedding approach for QA that encodes the links between KBs while also preserving the private feature of each KB by separating their embedding vectors. Nonetheless, by concurrently encoding the full and partial links, Multi-KB further surpasses Full-Link significantly, exemplifying the importance of fully comprehending all generalized links for an accurate Multi-KB-QA system. In this respect, despite the impressive results in conventional KB-QA, prior methods that omit the partial links may endure a poor performance upper bound in Multi-KB-QA. Hence, we believe that to a large extent, Multi-KB-QA remain unsolved by previous literatures, which justifies the formulation of this novel task and calls for enhancements on our approach for the future.

\subsection{Performance Breakdown}
To further validate the effectiveness of our method and the value of our task formulation argued in Section \ref{sec:overall_exp}, we breakdown the performance of each method on different question types in Table \ref{tab:breakdown_eval}. Although the overall performance of Full-Link and Merge-KB are close, Merge-KB mainly excels in questions with multi-hop reasoning (T1, T3, T6) that demand the system to grasp the strict equality between entities connected by a full link in the query graph, which naturally favors Merge-KB that directly fuses fully-linked entities. Besides, Full-Link stands out notably in T2 and T5 that require retrieving the common results of the one-hop reasoning on both KBs. In this scenario, Full-Link precisely encodes both the link relations and the private semantics of each KB to surpass Merge-KB, upholding the merit of our scheme to encode KBs independently without fusion.

Moreover, the superior performance of Multi-KB indeed mainly comes from its excellence on partial link questions, complying with its attempt to encode partial links in embedding. This substantiates the value of incorporating generalized links into normal KB-QA to formulate Multi-KB-QA.

\begin{table}[t]
\centering
\resizebox{.82\columnwidth}{!}{
\begin{tabular}{l l c c}
    \toprule
    \bfseries Method & \bfseries Type & \bfseries MRR & \bfseries Hits@1\\
    \cmidrule(lr){1-4}
    \multirow{2}{*}{\bfseries Plug-in} &
    KB1\textleftarrow KB2  & .438 & .349 \\
    & KB2\textleftarrow KB1  & .218 & .153  \\
    \cmidrule(lr){1-4}
    \multirow{2}{*}{\bfseries Plug-out} &
    -KB1  & .359 & .264  \\
    & -KB2 & .415 & .328  \\
    \bottomrule
\end{tabular}}
\caption{Pluggability performance on dev set. \textbf{Plug-in} refers to plugging a new KB to another well-trained KB in training (e.g.\textbf{KB1\textleftarrow KB2} means plugging in KB2). \textbf{Plug-out} denotes excluding a KB from candidate answers at inference (e.g.\textbf{-KB1} means unplugging KB1)}
\label{tab:plug}
\end{table}

\subsection{Pluggability of Multi-KB-QA}
As in Section \ref{sec:pluggability}, our embedding approach keep the pluggability to dynamically plug or unplug KBs, we present such ability in Table \ref{tab:plug}. In training, when the embedding of KB1 is obtained already, plugging in KB2 reaches competitive results to beat all baselines in Table \ref{tab:overall_eval} while requiring 90\% re-training cost (scale of training triples) compared to fully tuning both KBs; meanwhile, plugging KB1 to KB2 requires only 10\% re-training cost to reach reasonable performance, which largely boosts the flexibility of our system in practice. At inference, unplugging KB2 remains competitive while reducing 96\% candidate entity space, again proving the merit of pluggability. Due to the setting that all KBs are generally linked but not merged, such pluggability is not limited to our approach, but inherent to Multi-KB-QA, which is especially friendly for distributed data managing and federated learning to extend the value of this task.

\section{Related Works}
Typical KB-QA leverage a single open-domain \citep{lai2016open} or domain-specific \citep{haussmann2019foodkg} KB to answer one-hop factoid questions \citep{bordes2015large} or complex questions  that reason over multiple facts triples \citep{hu2018state, chen2021outlining}. To include more wide-ranging knowledge into KB-QA systems, works are done to answer complex question by multiple knowledge sources, e.g. to combine a KB with text corpus \citep{xu-etal-2016-question, sun2019pullnet, chen2020improving, lv2020graph}, structured tables \citep{oguz2020unik}, images \citep{shah2019kvqa}, or videos \citep{garcia2020knowit}. Differed from above, we embark on using multiple KBs for QA. Prior to us, \citet{lopez2012poweraqua} extracts fact triples relevant to the query intention from multi-KBs via linguistic rules and KB ontology matching, these triples are integrated to rank and select answers; \citet{fader2014open} splits a complex question into multiple one-hop queries by predefined linguistic patterns and mines templates to match each query to its answer in a specific KB. These works view multi-KBs as separate information sources while neglecting the inter-KB links, limiting their versatility in exploiting multi-KBs. Therefore, \citet{zhang2016joint} jointly discovers the entity alignments to merge KBs (by name similarity) and the triple patterns in the question (via entity and relation extraction) to inference answers; \citet{luo2022mu} merges the basic KB with an assisting KB by predefined alignments to enhance KB embedding and improve QA performance. In these studies, only the full links between KBs are considered and the KB fusion potentially loses the private feature and pluggability of each KB, we overcome these issues by formulating generalized links while preserving the independence of each KB in embedding.

Another separate line of research focus on adapting KB embedding for graph linking \citep{trivedi-etal-2018-linknbed}, logical reasoning \citep{ren2020query2box} or KB-QA \citep{saxena2020improving}, motivating us to encode generalized links in embedding for Multi-KB-QA.

\section{Conclusion}
We formulate the Multi-KB-QA task to spotlight the value of generalized links for QA on multi-KBs. By building the MKBQA benchmark and implementing our baseline method for Multi-KB-QA, we demonstrate that conventional KB-QA systems fail to perform well in Multi-KB-QA, justifying the task formulation and calling for the necessity of comprehending generalized links over multi-KBs for a robust KB-QA system. For the future, we expect to extend the scale and diversity of MKBQA to facilitate a more comprehensive evaluation for Multi-KB-QA.

\section*{Limitations}
\paragraph{Scale of Dataset} One of the primary aim of this paper is to emphasize the value of generalized links (not only full links) over multiple KBs for QA, we hence construct the novel MKBQA benchmark and verify by experiment that conventional KB-QA methods cannot successfully comprehend generalized links to fully unleash the power of multi-KBs in KB-QA. Due to the limitation of time and annotation cost, we set the scale of our dataset to $\sim$2.3k, which is still far from a massive scale KB-QA benchmark that covers more extensive generalized links and query intentions. By detailed experiment and analysis, we are convinced by the significance of formulating Multi-KB-QA and the effectiveness of our approach, calling for more subsequent studies to develop betters systems to leverage multiple generally-linked KBs. However, we are open to scale up the dataset for the future by discovering richer types of partial links and improving entity coverage, this might shape MKBQA as a more persuasive and sensitive benchmark for Multi-KB-QA.

\paragraph{Multi-KB-QA Approach Design} As argued in Section \ref{sec:method}, the KB embedding and QA method we propose in this paper is expected to be a simple yet effective baseline for Multi-KB-QA. Thus, straightforward enhancements could be made to further boost its performance (e.g. handling various partial links by a more fine-grained and delicate translation module). However, the experiments already demonstrate the superiority of our baseline over conventional KB-QA methods, which sufficiently establishes the value of the Multi-KB-QA task and supports the arguments of this paper. From this perspective, we leave the polishments for our approach as future works.

\section*{Ethics Statement}
The potential bias and mistake of our work may come from our dataset and KB-QA system. For instance, the query graphs in our dataset are based on the entities with generalized links, hence, the selection of knowledge facts in the dataset might be biased due to the fact that not all entities are connected by a full or partial link. Besides, since the background KB is curated from the internet, the knowledge in the KB might be erroneous, deviating the answers corresponding to the query graph from the actual answer in reality. However, this does not affect the fact that the query graph (and hence the question) correctly corresponds to the answers in the scope of our KB and remain the correctness of our benchmark. Finally, our KB-QA system may also generate false or biased answers when selecting entities from the KB given the nature of deep learning models, such answers is not the intention of our work (to formulate the Multi-KB-QA task and prove its effectiveness) and does not reflect our perspectives.


\bibliography{custom}

\begin{thebibliography}{39}
\expandafter\ifx\csname natexlab\endcsname\relax\def\natexlab#1{#1}\fi

\bibitem[{Balazevic et~al.(2019)Balazevic, Allen, and Hospedales}]{balazevic-etal-2019-tucker}
Ivana Balazevic, Carl Allen, and Timothy Hospedales. 2019.
\newblock \href {https://doi.org/10.18653/v1/D19-1522} {{T}uck{ER}: Tensor factorization for knowledge graph completion}.
\newblock In \emph{Proceedings of the 2019 Conference on Empirical Methods in Natural Language Processing and the 9th International Joint Conference on Natural Language Processing (EMNLP-IJCNLP)}, pages 5185--5194, Hong Kong, China. Association for Computational Linguistics.

\bibitem[{Berant et~al.(2013)Berant, Chou, Frostig, and Liang}]{berant2013semantic}
Jonathan Berant, Andrew Chou, Roy Frostig, and Percy Liang. 2013.
\newblock \href {https://aclanthology.org/D13-1160} {Semantic parsing on {F}reebase from question-answer pairs}.
\newblock In \emph{Proceedings of the 2013 Conference on Empirical Methods in Natural Language Processing}, pages 1533--1544, Seattle, Washington, USA. Association for Computational Linguistics.

\bibitem[{Bordes et~al.(2015)Bordes, Usunier, Chopra, and Weston}]{bordes2015large}
Antoine Bordes, Nicolas Usunier, Sumit Chopra, and Jason Weston. 2015.
\newblock Large-scale simple question answering with memory networks.
\newblock \emph{arXiv preprint arXiv:1506.02075}.

\bibitem[{Bordes et~al.(2013)Bordes, Usunier, Garc{\'{\i}}a{-}Dur{\'{a}}n, Weston, and Yakhnenko}]{Bordes2013TranslatingEF}
Antoine Bordes, Nicolas Usunier, Alberto Garc{\'{\i}}a{-}Dur{\'{a}}n, Jason Weston, and Oksana Yakhnenko. 2013.
\newblock \href {https://proceedings.neurips.cc/paper/2013/hash/1cecc7a77928ca8133fa24680a88d2f9-Abstract.html} {Translating embeddings for modeling multi-relational data}.
\newblock In \emph{Advances in Neural Information Processing Systems 26: 27th Annual Conference on Neural Information Processing Systems 2013. Proceedings of a meeting held December 5-8, 2013, Lake Tahoe, Nevada, United States}, pages 2787--2795.

\bibitem[{Chen et~al.(2020)Chen, Ji, Chen, and Zhang}]{chen2020improving}
Qianglong Chen, Feng Ji, Haiqing Chen, and Yin Zhang. 2020.
\newblock Improving commonsense question answering by graph-based iterative retrieval over multiple knowledge sources.
\newblock In \emph{Proceedings of the 28th International Conference on Computational Linguistics}, pages 2583--2594.

\bibitem[{Chen et~al.(2021)Chen, Li, Qi, Wu, and Wang}]{chen2021outlining}
Yongrui Chen, Huiying Li, Guilin Qi, Tianxing Wu, and Tenggou Wang. 2021.
\newblock Outlining and filling: Hierarchical query graph generation for answering complex questions over knowledge graph.
\newblock \emph{arXiv preprint arXiv:2111.00732}.

\bibitem[{Craswell(2009)}]{Craswell2009MeanRR}
Nick Craswell. 2009.
\newblock Mean reciprocal rank.
\newblock In \emph{Encyclopedia of Database Systems}.

\bibitem[{Cui et~al.(2021)Cui, Che, Liu, Qin, and Yang}]{cui-etal-2021-pretrain}
Yiming Cui, Wanxiang Che, Ting Liu, Bing Qin, and Ziqing Yang. 2021.
\newblock \href {https://doi.org/10.1109/TASLP.2021.3124365} {Pre-training with whole word masking for chinese bert}.

\bibitem[{Fader et~al.(2014)Fader, Zettlemoyer, and Etzioni}]{fader2014open}
Anthony Fader, Luke Zettlemoyer, and Oren Etzioni. 2014.
\newblock \href {https://doi.org/10.1145/2623330.2623677} {Open question answering over curated and extracted knowledge bases}.
\newblock In \emph{The 20th {ACM} {SIGKDD} International Conference on Knowledge Discovery and Data Mining, {KDD} '14, New York, NY, {USA} - August 24 - 27, 2014}, pages 1156--1165. {ACM}.

\bibitem[{Garcia et~al.(2020)Garcia, Otani, Chu, and Nakashima}]{garcia2020knowit}
Noa Garcia, Mayu Otani, Chenhui Chu, and Yuta Nakashima. 2020.
\newblock \href {https://aaai.org/ojs/index.php/AAAI/article/view/6713} {Knowit {VQA:} answering knowledge-based questions about videos}.
\newblock In \emph{The Thirty-Fourth {AAAI} Conference on Artificial Intelligence, {AAAI} 2020, The Thirty-Second Innovative Applications of Artificial Intelligence Conference, {IAAI} 2020, The Tenth {AAAI} Symposium on Educational Advances in Artificial Intelligence, {EAAI} 2020, New York, NY, USA, February 7-12, 2020}, pages 10826--10834. {AAAI} Press.

\bibitem[{Gu et~al.(2021)Gu, Kase, Vanni, Sadler, Liang, Yan, and Su}]{gu2021beyond}
Yu~Gu, Sue Kase, Michelle Vanni, Brian Sadler, Percy Liang, Xifeng Yan, and Yu~Su. 2021.
\newblock Beyond iid: three levels of generalization for question answering on knowledge bases.
\newblock In \emph{Proceedings of the Web Conference 2021}, pages 3477--3488.

\bibitem[{Haussmann et~al.(2019)Haussmann, Chen, Seneviratne, Rastogi, Codella, Chen, McGuinness, and Zaki}]{haussmann2019foodkg}
Steven Haussmann, Yu~Chen, Oshani Seneviratne, Nidhi Rastogi, James Codella, Ching-Hua Chen, Deborah~L McGuinness, and Mohammed~J Zaki. 2019.
\newblock Foodkg enabled q\&a application.
\newblock In \emph{ISWC Satellite Tracks}. CEUR-WS.

\bibitem[{Hu et~al.(2018)Hu, Zou, and Zhang}]{hu2018state}
Sen Hu, Lei Zou, and Xinbo Zhang. 2018.
\newblock \href {https://doi.org/10.18653/v1/D18-1234} {A state-transition framework to answer complex questions over knowledge base}.
\newblock In \emph{Proceedings of the 2018 Conference on Empirical Methods in Natural Language Processing}, pages 2098--2108, Brussels, Belgium. Association for Computational Linguistics.

\bibitem[{Kv{\aa}lseth(1989)}]{kvaalseth1989note}
Tarald~O Kv{\aa}lseth. 1989.
\newblock Note on cohen's kappa.
\newblock \emph{Psychological reports}, 65(1):223--226.

\bibitem[{Lai et~al.(2016)Lai, Lin, Chen, Feng, and Zhao}]{lai2016open}
Yuxuan Lai, Yang Lin, Jiahao Chen, Yansong Feng, and Dongyan Zhao. 2016.
\newblock Open domain question answering system based on knowledge base.
\newblock In \emph{Natural Language Understanding and Intelligent Applications}, pages 722--733. Springer.

\bibitem[{Larr{\'e}ch{\'e} and Moinpour(1983)}]{Larrch1983ManagerialJI}
Jean-Claude Larr{\'e}ch{\'e} and Reza Moinpour. 1983.
\newblock Managerial judgment in marketing: The concept of expertise.
\newblock \emph{Journal of Marketing Research}, 20:110 -- 121.

\bibitem[{Liu et~al.(2019)Liu, Ott, Goyal, Du, Joshi, Chen, Levy, Lewis, Zettlemoyer, and Stoyanov}]{Liu2019RoBERTaAR}
Yinhan Liu, Myle Ott, Naman Goyal, Jingfei Du, Mandar Joshi, Danqi Chen, Omer Levy, Mike Lewis, Luke Zettlemoyer, and Veselin Stoyanov. 2019.
\newblock Roberta: A robustly optimized bert pretraining approach.
\newblock \emph{ArXiv}, abs/1907.11692.

\bibitem[{Lopez et~al.(2012)Lopez, Fern{\'a}ndez, Motta, and Stieler}]{lopez2012poweraqua}
Vanessa Lopez, Miriam Fern{\'a}ndez, Enrico Motta, and Nico Stieler. 2012.
\newblock Poweraqua: Supporting users in querying and exploring the semantic web.
\newblock \emph{Semantic web}, 3(3):249--265.

\bibitem[{Luo et~al.(2022)Luo, Sun, and Hu}]{luo2022mu}
Xindi Luo, Zequn Sun, and Wei Hu. 2022.
\newblock $\mu$kg: A library for multi-source knowledge graph embeddings and applications.
\newblock In \emph{International Semantic Web Conference}, pages 610--627. Springer.

\bibitem[{Lv et~al.(2020)Lv, Guo, Xu, Tang, Duan, Gong, Shou, Jiang, Cao, and Hu}]{lv2020graph}
Shangwen Lv, Daya Guo, Jingjing Xu, Duyu Tang, Nan Duan, Ming Gong, Linjun Shou, Daxin Jiang, Guihong Cao, and Songlin Hu. 2020.
\newblock Graph-based reasoning over heterogeneous external knowledge for commonsense question answering.
\newblock In \emph{Proceedings of the AAAI conference on artificial intelligence}, volume~34, pages 8449--8456.

\bibitem[{Ngomo(2018)}]{ngomo20189th}
Ngonga Ngomo. 2018.
\newblock 9th challenge on question answering over linked data (qald-9).
\newblock \emph{language}, 7(1):58--64.

\bibitem[{Oguz et~al.(2022)Oguz, Chen, Karpukhin, Peshterliev, Okhonko, Schlichtkrull, Gupta, Mehdad, and Yih}]{oguz2020unik}
Barlas Oguz, Xilun Chen, Vladimir Karpukhin, Stan Peshterliev, Dmytro Okhonko, Michael Schlichtkrull, Sonal Gupta, Yashar Mehdad, and Scott Yih. 2022.
\newblock \href {https://doi.org/10.18653/v1/2022.findings-naacl.115} {{U}ni{K}-{QA}: Unified representations of structured and unstructured knowledge for open-domain question answering}.
\newblock In \emph{Findings of the Association for Computational Linguistics: NAACL 2022}, pages 1535--1546, Seattle, United States. Association for Computational Linguistics.

\bibitem[{Petrochuk and Zettlemoyer(2018)}]{petrochuk-zettlemoyer-2018-simplequestions}
Michael Petrochuk and Luke Zettlemoyer. 2018.
\newblock \href {https://doi.org/10.18653/v1/D18-1051} {{S}imple{Q}uestions nearly solved: A new upperbound and baseline approach}.
\newblock In \emph{Proceedings of the 2018 Conference on Empirical Methods in Natural Language Processing}, pages 554--558, Brussels, Belgium. Association for Computational Linguistics.

\bibitem[{Ren et~al.(2020)Ren, Hu, and Leskovec}]{ren2020query2box}
Hongyu Ren, Weihua Hu, and Jure Leskovec. 2020.
\newblock \href {https://openreview.net/forum?id=BJgr4kSFDS} {Query2box: Reasoning over knowledge graphs in vector space using box embeddings}.
\newblock In \emph{8th International Conference on Learning Representations, {ICLR} 2020, Addis Ababa, Ethiopia, April 26-30, 2020}. OpenReview.net.

\bibitem[{Saxena et~al.(2020)Saxena, Tripathi, and Talukdar}]{saxena2020improving}
Apoorv Saxena, Aditay Tripathi, and Partha Talukdar. 2020.
\newblock \href {https://doi.org/10.18653/v1/2020.acl-main.412} {Improving multi-hop question answering over knowledge graphs using knowledge base embeddings}.
\newblock In \emph{Proceedings of the 58th Annual Meeting of the Association for Computational Linguistics}, pages 4498--4507, Online. Association for Computational Linguistics.

\bibitem[{Shah et~al.(2019)Shah, Mishra, Yadati, and Talukdar}]{shah2019kvqa}
Sanket Shah, Anand Mishra, Naganand Yadati, and Partha~Pratim Talukdar. 2019.
\newblock \href {https://doi.org/10.1609/aaai.v33i01.33018876} {{KVQA:} knowledge-aware visual question answering}.
\newblock In \emph{The Thirty-Third {AAAI} Conference on Artificial Intelligence, {AAAI} 2019, The Thirty-First Innovative Applications of Artificial Intelligence Conference, {IAAI} 2019, The Ninth {AAAI} Symposium on Educational Advances in Artificial Intelligence, {EAAI} 2019, Honolulu, Hawaii, USA, January 27 - February 1, 2019}, pages 8876--8884. {AAAI} Press.

\bibitem[{Song et~al.(2015)Song, Schilder, Smiley, Brew, Zielund, Bretz, Martin, Dale, Duprey, Miller et~al.}]{song2015tr}
Dezhao Song, Frank Schilder, Charese Smiley, Chris Brew, Tom Zielund, Hiroko Bretz, Robert Martin, Chris Dale, John Duprey, Tim Miller, et~al. 2015.
\newblock Tr discover: A natural language interface for querying and analyzing interlinked datasets.
\newblock In \emph{International Semantic Web Conference}, pages 21--37. Springer.

\bibitem[{Soukoreff and MacKenzie(2001)}]{Soukoreff2001MeasuringEI}
R.~William Soukoreff and Ian~Scott MacKenzie. 2001.
\newblock Measuring errors in text entry tasks: an application of the levenshtein string distance statistic.
\newblock \emph{CHI '01 Extended Abstracts on Human Factors in Computing Systems}.

\bibitem[{Sun et~al.(2019)Sun, Bedrax-Weiss, and Cohen}]{sun2019pullnet}
Haitian Sun, Tania Bedrax-Weiss, and William Cohen. 2019.
\newblock \href {https://doi.org/10.18653/v1/D19-1242} {{P}ull{N}et: Open domain question answering with iterative retrieval on knowledge bases and text}.
\newblock In \emph{Proceedings of the 2019 Conference on Empirical Methods in Natural Language Processing and the 9th International Joint Conference on Natural Language Processing (EMNLP-IJCNLP)}, pages 2380--2390, Hong Kong, China. Association for Computational Linguistics.

\bibitem[{Trivedi et~al.(2017)Trivedi, Maheshwari, Dubey, and Lehmann}]{trivedi2017lc}
Priyansh Trivedi, Gaurav Maheshwari, Mohnish Dubey, and Jens Lehmann. 2017.
\newblock Lc-quad: A corpus for complex question answering over knowledge graphs.
\newblock In \emph{International Semantic Web Conference}, pages 210--218. Springer.

\bibitem[{Trivedi et~al.(2018)Trivedi, Sisman, Dong, Faloutsos, Ma, and Zha}]{trivedi-etal-2018-linknbed}
Rakshit Trivedi, Bunyamin Sisman, Xin~Luna Dong, Christos Faloutsos, Jun Ma, and Hongyuan Zha. 2018.
\newblock \href {https://doi.org/10.18653/v1/P18-1024} {{L}ink{NB}ed: Multi-graph representation learning with entity linkage}.
\newblock In \emph{Proceedings of the 56th Annual Meeting of the Association for Computational Linguistics (Volume 1: Long Papers)}, pages 252--262, Melbourne, Australia. Association for Computational Linguistics.

\bibitem[{Trouillon et~al.(2016)Trouillon, Welbl, Riedel, Gaussier, and Bouchard}]{Trouillon2016ComplexEF}
Th{\'{e}}o Trouillon, Johannes Welbl, Sebastian Riedel, {\'{E}}ric Gaussier, and Guillaume Bouchard. 2016.
\newblock \href {http://proceedings.mlr.press/v48/trouillon16.html} {Complex embeddings for simple link prediction}.
\newblock In \emph{Proceedings of the 33nd International Conference on Machine Learning, {ICML} 2016, New York City, NY, USA, June 19-24, 2016}, volume~48 of \emph{{JMLR} Workshop and Conference Proceedings}, pages 2071--2080. JMLR.org.

\bibitem[{Unger et~al.(2012)Unger, B{\"{u}}hmann, Lehmann, Ngomo, Gerber, and Cimiano}]{unger2012template}
Christina Unger, Lorenz B{\"{u}}hmann, Jens Lehmann, Axel{-}Cyrille~Ngonga Ngomo, Daniel Gerber, and Philipp Cimiano. 2012.
\newblock \href {https://doi.org/10.1145/2187836.2187923} {Template-based question answering over {RDF} data}.
\newblock In \emph{Proceedings of the 21st World Wide Web Conference 2012, {WWW} 2012, Lyon, France, April 16-20, 2012}, pages 639--648. {ACM}.

\bibitem[{Wang et~al.(2015)Wang, Berant, and Liang}]{Wang2015BuildingAS}
Yushi Wang, Jonathan Berant, and Percy Liang. 2015.
\newblock \href {https://doi.org/10.3115/v1/P15-1129} {Building a semantic parser overnight}.
\newblock In \emph{Proceedings of the 53rd Annual Meeting of the Association for Computational Linguistics and the 7th International Joint Conference on Natural Language Processing (Volume 1: Long Papers)}, pages 1332--1342, Beijing, China. Association for Computational Linguistics.

\bibitem[{Xu et~al.(2016)Xu, Reddy, Feng, Huang, and Zhao}]{xu-etal-2016-question}
Kun Xu, Siva Reddy, Yansong Feng, Songfang Huang, and Dongyan Zhao. 2016.
\newblock \href {https://doi.org/10.18653/v1/P16-1220} {Question answering on {F}reebase via relation extraction and textual evidence}.
\newblock In \emph{Proceedings of the 54th Annual Meeting of the Association for Computational Linguistics (Volume 1: Long Papers)}, pages 2326--2336, Berlin, Germany. Association for Computational Linguistics.

\bibitem[{Yih et~al.(2016)Yih, Richardson, Meek, Chang, and Suh}]{yih2016value}
Wen-tau Yih, Matthew Richardson, Chris Meek, Ming-Wei Chang, and Jina Suh. 2016.
\newblock \href {https://doi.org/10.18653/v1/P16-2033} {The value of semantic parse labeling for knowledge base question answering}.
\newblock In \emph{Proceedings of the 54th Annual Meeting of the Association for Computational Linguistics (Volume 2: Short Papers)}, pages 201--206, Berlin, Germany. Association for Computational Linguistics.

\bibitem[{Zeng et~al.(2021)Zeng, Li, Hou, Li, and Feng}]{zeng2021comprehensive}
Kaisheng Zeng, Chengjiang Li, Lei Hou, Juanzi Li, and Ling Feng. 2021.
\newblock A comprehensive survey of entity alignment for knowledge graphs.
\newblock \emph{AI Open}, 2:1--13.

\bibitem[{Zhang et~al.(2022)Zhang, Zhang, Li, and Zou}]{zhang2022crake}
Minhao Zhang, Ruoyu Zhang, Yanzeng Li, and Lei Zou. 2022.
\newblock \href {https://doi.org/10.18653/v1/2022.findings-naacl.136} {Crake: Causal-enhanced table-filler for question answering over large scale knowledge base}.
\newblock In \emph{Findings of the Association for Computational Linguistics: NAACL 2022}, pages 1787--1798, Seattle, United States. Association for Computational Linguistics.

\bibitem[{Zhang et~al.(2016)Zhang, He, Liu, and Zhao}]{zhang2016joint}
Yuanzhe Zhang, Shizhu He, Kang Liu, and Jun Zhao. 2016.
\newblock \href {http://www.aaai.org/ocs/index.php/AAAI/AAAI16/paper/view/12080} {A joint model for question answering over multiple knowledge bases}.
\newblock In \emph{Proceedings of the Thirtieth {AAAI} Conference on Artificial Intelligence, February 12-17, 2016, Phoenix, Arizona, {USA}}, pages 3094--3100. {AAAI} Press.

\end{thebibliography}
\bibliographystyle{acl_natbib}

\appendix

\begin{figure*}[t]
    \centering
    \includegraphics[width=0.84\linewidth]{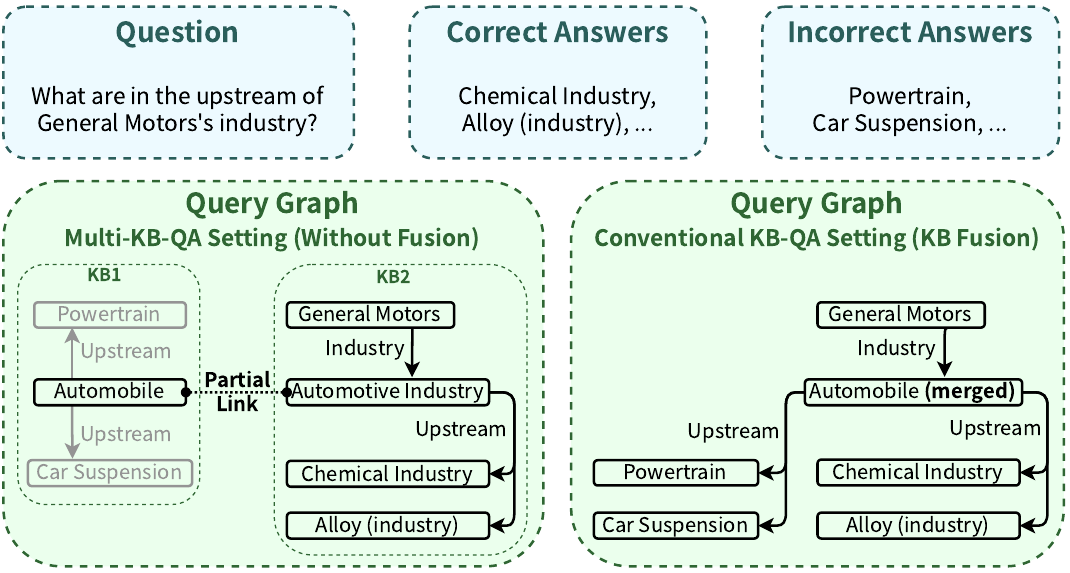}
    \caption{A failure case when forcibly merging partially-linked entities. The question only queries the facts in KB2 (i.e. the upstream industries of \textit{Automotive Industry}). However, forcibly fusing the partial link connects both the upstream products of \textit{Automobile} and the upstream industries of \textit{Automotive Industry} to the merged entity \textit{Automobile (merged)}. Consequently, a QA system running on the merged KB cannot distinguish upstream industries (correct answers) from upstream products (incorrect answers), resulting in a low QA precision. Besides, the Multi-KB-QA setting sidesteps such issue by remaining each KB separate. A robust Multi-KB-QA system can grasp the inter-KB link relations and select the correct answers. }
    \label{fig:merge_failure}
\end{figure*}

\section{Discussion of Entity Alignment on Partial Links}
\label{sec:partial_failure}

To utilize multiple KBs in answering complex questions, former studies fuse all KBs into a single KB via entity alignment and reduce the task into conventional single-KB-QA \citep{zhang2016joint, luo2022mu}. As explained in Section \ref{sec:intro} and Figure \ref{fig:full_partial_link}, however, such fashion appears inadequate in fully representing the partial links between KBs. In this section, we discuss such deficiency of KB fusion over partial links in more detail.

Conceptually, a partial link exists between a pair of entities referring to different facts (Section \ref{sec:form_gen_links}), while entity alignment discovers entity pairs that refer to the same object. In this regard, previous studies based on KB-fusion cannot align partial links and may therefore fail in answering complex queries involving partial links (e.g. the question in Figure \ref{fig:full_partial_link}-right). 

Furthermore, even if we forcibly merge the entity pairs with partial links in KB-fusion, the semantics in the KBs might be corrupted, again resulting in a degraded QA performance. Consider the example in Figure \ref{fig:merge_failure}, according to the original ontology in each KB, the downstream of products and industries are of different types, but merging the partial link connects both products and industries to a same entity. Thus, forcibly merging the partially-linked entities corrupts the KB semantics, which confuses a QA system in constructing query graphs and selecting accurate answers.

On the contrary, the Multi-KB-QA setting leaves different KBs separate to preserve the rich semantics included in the generalized links among KBs. From this perspective, such setting is capable of handling more general cases (involving both full and partial links) in practice comparing with the former KB-fusion frameworks. We believe that this justifies the significance of devising the Multi-KB-QA task in this work.

\begin{figure*}[h!]
    \centering
    \includegraphics[width=0.99\linewidth]{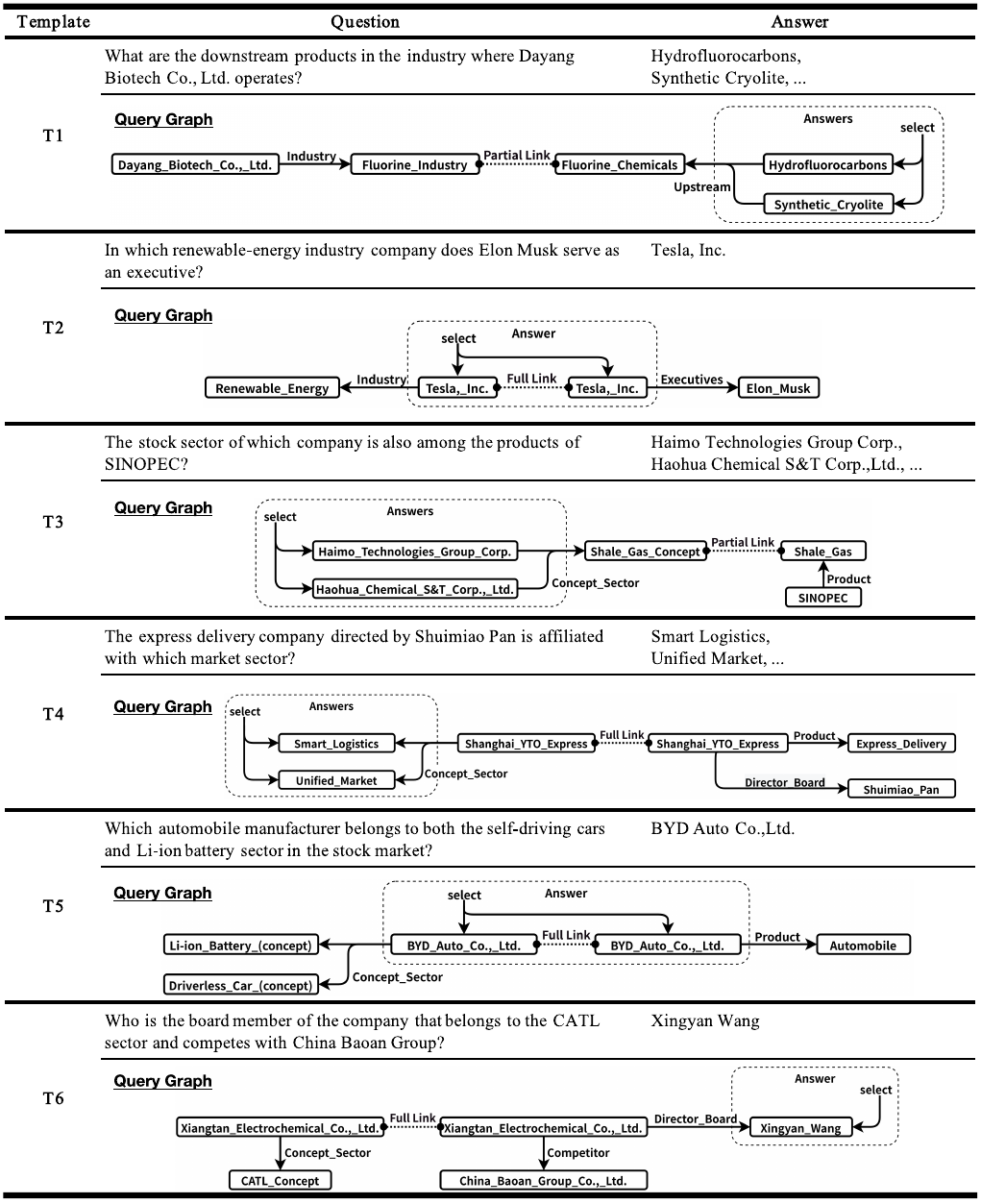}
    \caption{ Example query graphs instantiated from each template defined in Figure \ref{fig:query_graph}. }
    \label{fig:example_data}
\end{figure*}

\section{Annotating the MKBQA Dataset}
\label{sec:appendix_anno}
\subsection{Annotation Setup}
Following the annotation framework discussed in Section \ref{sec:construction_dataset}, we recruit 5 annotators (with 4 graduate and 1 undergraduate students) to paraphrase the canonical questions into natural and diversified questions as the input of Multi-KB-QA models. Since a high-quality paraphrase requires both a precise comprehension of the query graph over multiple KBs and the ability to generate syntactically diversified questions for KB-QA, normal crowdsourcing process may not be fit the demands. Hence, we adopt the managerial judgement \citep{Larrch1983ManagerialJI} framework to recruit annotators with sufficient prior experience in constructing conventional Chinese KB-QA dataset or implementing KB-QA systems. Further, we provide a detailed guideline including case analysis to prompt precise and diversified paraphrase, the annotation task is dispatched only when we ensure each annotator thoroughly grasps the guideline and understands the usage and potential risk of the dataset. We provide essential payment when each annotator finishes the task.

\subsection{Agreement Between Annotators}
Unlike many classification-styled annotation task, we do not apply metrics like Cohen's Kappa \citep{kvaalseth1989note} to measure the agreement between annotators. Since the paraphrasing involved is a typical natural language generation task and has no "gold paraphrase", it's hard to measure the extent to which annotations agree with each other or with the correct question intention. Besides, we actually hope to bring about more linguistic diversity to the dataset via paraphasing. In this regard, we also prefer not to let annotators stick to any specific paraphrasing format (while potentially reducing text-similarity metrics like BLEU).

To ensure the agreement between all reviewers, as introduced in Section \ref{sec:construction_dataset}, we proceed the annotation process in a two-staged manner. That is, we first annotate 5\% of the data and let each annotator to come across all these paraphrases to discuss and report any possible mistakes or issues in the paraphrase. In this process, 6 out of 100 annotations were reported as incorrect (e.g. misused some facts/predicates in the question). We feedback these mistakes to all annotators to ensure that they understand how to write a "proper" paraphrase. With this first annotation stage, we believe that all annotators reached sufficient agreement to proceed onto the rest of the dataset. 

\subsection{Potential Risks of the Dataset}
To avoid the potential leak of private information or offensive content in the question, we manually checked 10\% of the paraphrased questions and found no sensitive content. Further, the KBs we use are extracted from the publicly available information on the internet (e.g. the announcements and reports found on the official website of the companies), no private or sensitive content is gathered in the construction of the KB. To this regard, we believe the MKBQA benchmark does not contain offensive or sensitive information.

\section{Examples of Query Template Instantiation}
\label{sec:example_data}

We provide example query graphs (with the corresponding question and answers) derived from each template in Figure \ref{fig:example_data}. For instance, the first template T1 in Figure \ref{fig:query_graph} is instantiated by the first query graph in Figure \ref{fig:example_data}. Specifically, the entity \textit{Dayang\_Biotech\_Co.,\_Ltd.} is bound to \textit{<e1>}, the relation \textit{Industry} and \textit{Downstream} are bound to \textit{<r1>} and \textit{<r2>} respectively. Hence, the variable \textit{?v1} is corresponded by \textit{Fluorine\_Industry} in KB1, which has a partial link with \textit{Fluorine\_Chemicals} in KB2, leading to the answers \textit{Hydrofluorocarbons} and \textit{Synthetic\_Cryolite} (i.e. the upstream products of fluorine chemicals) to correspond with \textit{?v2}. The similar process is applied to each template to build our MKBQA dataset.

\section{Hyperparameter Settings}
\label{sec:appendix_hyper}
The hyperparameters used to train our best model for Multi-KB-QA are displayed in Table \ref{tab:hyper}.

\begin{table}[h]
\centering
\resizebox{1.\columnwidth}{!}{
\begin{tabular}{c l c}
    \toprule
    \bfseries Name & \bfseries Description & \bfseries Setting\\
    \cmidrule(lr){1-3}
    $d$ & Hidden size of the RoBERTa encoder & 1024\\
    $h$ & Dimension of the embedding vector & 200\\
    $optim$ & Optimizer to train KB embedding and QA model & Adam\\
    $batch_{kbe}$ & Batch size in training the KB embedding & 128\\
    $n_{kbe}$ & Maximum epochs of KB embedding training & 400\\
    $n_{warm}$ & Warmup epochs to only train on replaced triples $R$ & 30\\
    $lr_{kbe}$ & Learning rate of the KB embedding vectors & 1e-3\\
    $lr_{trans}$ & Learning rate of the translator & 5e-4\\
    $k_{kbe}$ & Size of negative samples in KB embedding & 1000\\
    $k_{pl}$ & Times of duplication for partial link triples in $R$ & 10\\
    $\gamma_{kbe}$ & Label-smoothing ratio in KB embedding & 0.1\\
    $p_{kbe}$ & Dropout probability of the embedding vectors & 0.3\\
    $r_{lk}$ & Ratio of the loss on replaced triples $\ell_{link}$ & 2.25\\
    $batch_{qa}$ & Batch size of the QA model & 32\\
    $n_{kbe}$ & Maximum epochs of KB embedding training & 1000\\
    $lr_{qa}$ & Learning rate of the QA model & 1e-5\\
    $k_{qa}$ & Size of negative samples of the QA model & 500\\
    $\gamma_{qa}$ & Label-smoothing ratio in QA model training & 0.05\\
    \bottomrule
\end{tabular}}
\caption{Hyperparameter settings for our approach.}\label{tab:hyper}
\end{table}

\end{document}